\documentclass[lettersize,journal]{IEEEtran}
\usepackage{amsmath,amsfonts}
\usepackage{algorithmic}
\usepackage{algorithm}
\usepackage{array}
\usepackage[caption=false,font=normalsize,labelfont=sf,textfont=sf]{subfig}
\usepackage{textcomp}
\usepackage{stfloats}
\usepackage{url}
\usepackage{verbatim}
\usepackage{graphicx}
\usepackage{cite}
\usepackage{booktabs}
\usepackage{multirow}
\usepackage{bm}
\usepackage{tabularx}
\usepackage{rotating}
\usepackage{longtable}
\usepackage{booktabs}
\usepackage{csquotes}
\usepackage{multirow}
\usepackage{xcolor}

\begin{document}

\title{MoveFM-R: Advancing Mobility Foundation Models via Language-driven Semantic Reasoning}

\author{
    Fanjin Meng, Yuan Yuan, Jingtao Ding, Jie Feng, Chonghua Han, and Yong Li%
    \thanks{Fanjin Meng and Yuan Yuan contributed equally to this work.}%
    \thanks{Jingtao Ding and Yong Li are the corresponding authors.}%
    \thanks{Fanjin Meng, Jingtao Ding, Jie Feng, Chonghua Han, and Yong Li are with the Department of Electronic Engineering, BNRist, Tsinghua University, Beijing, China. (e-mail: mengfj23@tsinghua.edu.cn; dingjt15@tsinghua.org.cn; liyong07@tsinghua.edu.cn).}%
    \thanks{Yuan Yuan is with New York University (NYU). (e-mail: y-yuan20@tsinghua.org.cn).}%
}

\maketitle

\begin{abstract}
Mobility Foundation Models (MFMs) have significantly advanced the modeling of human movement patterns. However, they face a performance ceiling imposed by the lack of understanding of the latent semantic intent behind user behaviors. This restricts their ability to generate samples outside the training distribution, thereby limiting their effectiveness in complex real-world applications. A prime example is counterfactual reasoning: standard models struggle to simulate how trajectories would shift under novel constraints, such as a sudden traffic accident, because they rely on historical inertia. 
While Large Language Models (LLMs) offer powerful semantic reasoning, they lack the innate understanding of spatio-temporal statistics required for generating physically plausible mobility trajectories. 
To address these challenges, we propose MoveFM-R, a framework that synergizes the statistical fidelity of MFMs with the semantic reasoning of LLMs. MoveFM-R bridges the fundamental modality gap through a Semantically Enhanced Location Encoding and a curriculum that aligns MFM's latent spatial representations with the LLM's semantic space. Furthermore, we introduce a Self-Reflective Mechanism that iteratively edits trajectories to ensure compliance with counterfactual constraints while preserving realistic mobility laws. Extensive experiments demonstrate that MoveFM-R significantly outperforms baselines, showing robust generalization and superior capability in handling ``what-if" scenarios. 
\end{abstract}


\section{Introduction}

\IEEEPARstart{T}{he} proliferation of large-scale mobility data, such as GPS trajectories and location-based service records, has fueled the development of Mobility Foundation Models (MFMs)~\cite{zhou2024urban}. Unlike traditional task-specific models, MFMs are pre-trained on massive, multi-city datasets to learn universal movement patterns. 
By capturing the spatiotemporal transition dynamics common across urban environments, these models have demonstrated remarkable capabilities in trajectory prediction and generation, exhibiting improved generalization compared to traditional methods, even when applied to unseen cities or regions~\cite{zhu2024unitraj,han2025trajmoe,liu2024moirai,long2025one}.

Despite their advancements, MFMs encounter fundamental developmental bottlenecks imposed by the nature of their training data and learning objectives. 
First, MFMs primarily model the statistical transition probabilities between numerical coordinates. While they can predict where a user might go next based on historical frequency, they struggle to comprehend the semantic \textit{why}—the human intents and functional purposes driving these movements. 
Second, limited by privacy concerns and collection costs, the scale of available mobility data remains significantly smaller than the web-scale corpora used for Large Language Models (LLMs)~\cite{kim2020location}. 
These deficiencies in semantic grounding and data scale impose a performance ceiling on MFMs.
Third, MFMs face a capability deficit regarding counterfactual (\enquote{what-if}) generation\footnote{We frame this as conditional trajectory generation under novel constraints, rather than strictly mathematical causal inference.}. Constrained by factual historical data and lacking the semantic reasoning required for extrapolation, they fail to simulate scenarios that deviate from established distributions. For instance, if a traffic accident necessitates a detour, a standard MFM will likely continue to generate trajectories following historical habits, ignoring the novel external constraint.

The emergence of Large Language Models (LLMs) promises to address these bottlenecks.
Since 2024, researchers have begun exploring LLMs for mobility tasks~\cite{gong2024mobility,chen2025enhancing}, aiming to leverage the rich semantic knowledge and reasoning capabilities acquired during pre-training. 
Integrating LLMs has the potential to mitigate the inherent limitations of MFMs: their vast world knowledge and few-shot learning capabilities can compensate for the semantic scarcity and limited data scale of mobility records, while their logical reasoning abilities are naturally suited for handling \enquote{what-if} counterfactual questions that require extrapolating beyond historical statistics. 
However, simply replacing MFMs with standalone LLMs is insufficient. 
Without an innate understanding of spatiotemporal statistics for precise urban mobility modeling, pure LLMs often struggle to ground their reasoning in physical reality, generating semantically coherent but geographically incoherent sequences or physically infeasible paths~\cite{koda2025locationreasoner,shao2024chain,wang2024large}. 
The optimal path forward is therefore synthesis, not replacement.

Nevertheless, simply cascading an LLM with an MFM is not enough. To truly augment MFMs with LLM capabilities, we must overcome two primary challenges:
\textbf{1) The Comprehension Gap:} There is a fundamental disconnect between the two modalities. At the underlying location level, there is a vocabulary mismatch between continuous geographic coordinates and discrete language tokens~\cite{chen2025enhancing}. At the trajectory level, the latent vector representations of MFMs (optimized for spatial distance) are misaligned with the semantic representation space of LLMs~\cite{hashemi2025points}. Without bridging this gap, the LLM cannot effectively interpret the physical information captured by the MFM.
\textbf{2) The \enquote{what-if} Generation Dilemma:} Achieving counterfactual generation requires a delicate balance between fidelity and compliance. The model must satisfy novel, unseen spatiotemporal constraints (e.g., avoiding area X due to an accident) while still adhering to the general physical laws of human mobility and the user's historical habits. Pure LLMs tend to hallucinate impossible paths to satisfy the prompt, while pure MFMs ignore the constraints to satisfy historical statistics.

To address these challenges, we propose \textbf{MoveFM-R}, a novel framework that synergizes the statistical power of MFMs with the semantic reasoning of LLMs.
To bridge the comprehension gap, we introduce a Semantically Enhanced Location Encoding (a universal codebook) and a Progressive Alignment Curriculum. These modules translate raw coordinates into semantic tokens and guide the LLM to interpret MFM features, moving from low-level descriptions to high-level pattern summarization. 
To resolve the \enquote{what-if} dilemma, we draw inspiration from image editing tasks in computer vision, where a model modifies an image to match a prompt while preserving its original structure—and reframe counterfactual generation as a \enquote{trajectory editing} task. 
Instead of generating a trajectory from scratch, we introduce a Self-Reflective Mechanism based on Reinforcement Learning (RL). 
This mechanism allows the model to iteratively \enquote{edit} a baseline trajectory derived from user historical patterns, critiquing its own output to maximize compliance with language instructions while minimizing deviation from realistic mobility distributions.
In summary, our contributions are as follows:
\begin{itemize}
    \item We propose a new paradigm that synthesizes LLMs with MFMs, enhancing the statistical backbone of MFMs with the advanced semantic reasoning of LLMs to enable more comprehensive mobility modeling.
    \item We introduce MoveFM-R, which features a universal semantic codebook and an alignment curriculum to bridge the modality gap, alongside a self-reflective trajectory editing mechanism to enable robust what-if generation.
    \item We demonstrate state-of-the-art performance on mobility prediction and generation through extensive experiments, showing significant improvements over MFM and LLM baselines, robust zero-shot generalization, and the ability to generate counterfactual trajectories from natural language instructions.
\end{itemize}

\section{Related Work}

\subsection{Building mobility foundational models from scratch.}
The availability of large-scale trajectory data has facilitated the development of foundational models for human mobility. Early work, such as the Pretrained Mobility Transformer (PMT)~\cite{wu2024pretrained}, demonstrated that large-scale pre-training can capture transferable, region-independent movement patterns. Subsequent research has expanded this paradigm, including enhancing cross-city transfer capabilities~\cite{LIUKang1520}, exploring generative frameworks such as diffusion models~\cite{chu2023trajgdm}, and leveraging mixture-of-experts (MoE) architectures for improved scalability~\cite{zhu2024unitraj,liu2024moirai,shi2024time,han2025trajmoe,wei2025transfertraj}.
Despite their success in modeling statistical patterns, these models operate on coordinate and sequence-based language and lack inherent mechanisms for understanding high-level semantics and human intent. This fundamentally limits their reasoning capabilities and motivates the integration of LLMs.

\begin{figure*}[t]
    \centering
    \includegraphics[width=1.0\linewidth]{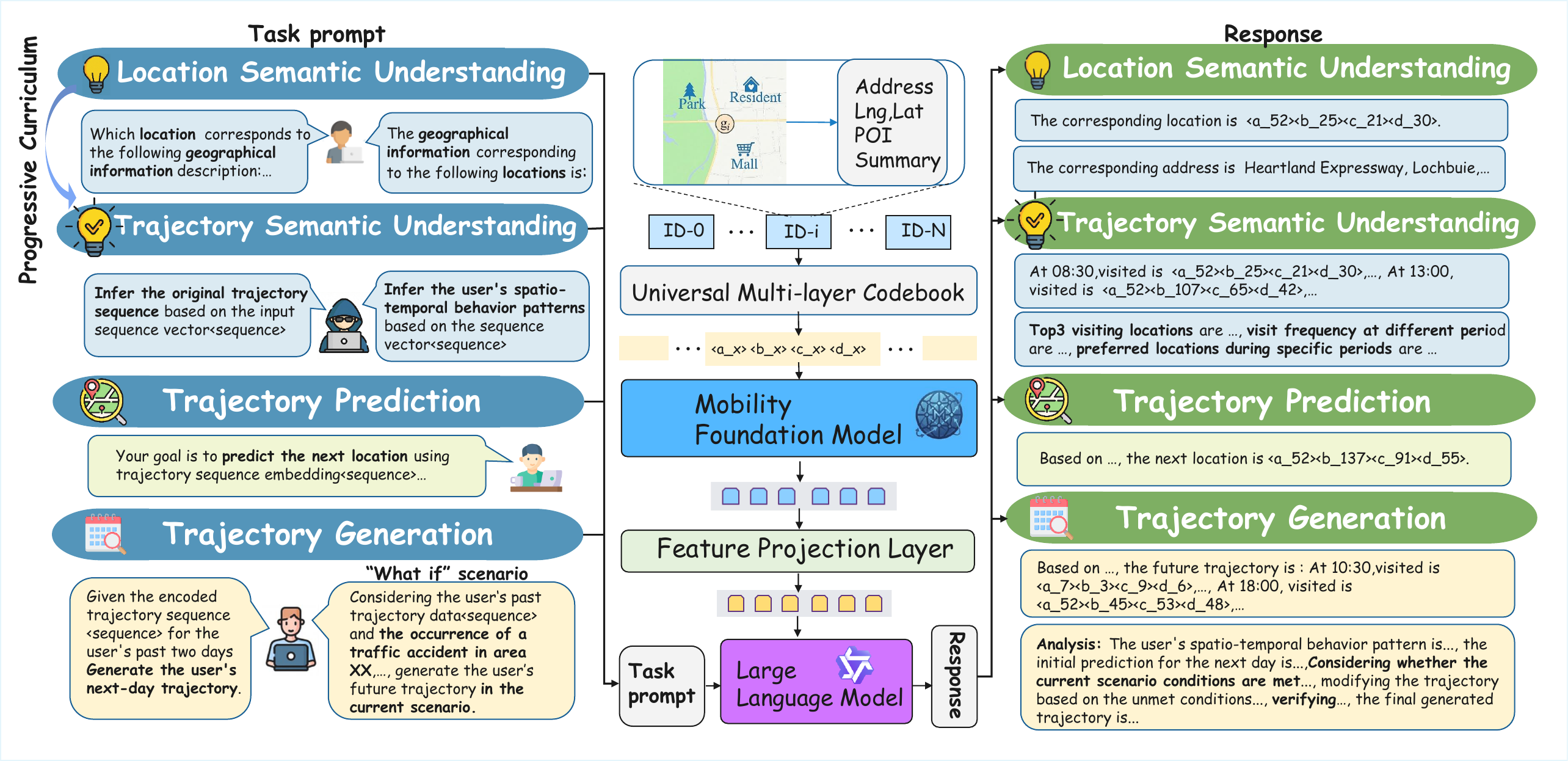}
    \vspace{-6mm}
    \caption{MoveFM-R Model Architecture.}
    \label{fig:framework}
\end{figure*} 

\subsection{LLM-Based Mobility Modeling}
Recently, researchers have explored the application of large language models (LLMs) to the mobility domain. Through specialized codebooks or sequence reprogramming~\cite{gong2024mobility,chib2024lg,chen2025enhancing,wang2025generative}, continuous trajectories are aligned with the discrete input space of the LLM. On the generative side, researchers have encouraged LLMs to simulate human decision-making processes~\cite{shao2024chain,wang2024simulating} or act as urban agents to generate trajectories~\cite{wang2024large,ju2025trajllm}. Another approach is to enrich the original trajectories with semantic attributes, such as points of interest (POIs) or activities, to improve model performance~\cite{luo2024deciphering,liu2024nextlocllm,lan2024traj}.
While these studies successfully incorporate semantic knowledge into mobility modeling, they must reduce continuous trajectories to discrete sequences of symbols in order to make spatiotemporal data digestible to LLMs. This process typically sacrifices geometric accuracy and can produce trajectories that are semantically plausible but geographically incoherent or physically unfeasible. 
We propose MoveFM-R to address this fundamental challenge. By synthesizing the semantic reasoning capabilities of LLMs with the statistical fidelity of a dedicated mobility encoder, our framework enables more comprehensive and physically constrained mobility modeling.

\section{Preliminaries}
\label{sec:preliminaries}

\subsection{Mobility Foundation Model (MFM)}
Conceptually, an MFM is a probabilistic model pre-trained on massive, multi-city human mobility data to learn universal movement patterns. Recent advancements have leveraged scalable architectures, such as Transformer-based sequence networks~\cite{wu2024pretrained} and Mixture-of-Experts (MoE)~\cite{zhu2024unitraj,han2025trajmoe}, to process billion-scale traces. By extensively modeling spatiotemporal transition dynamics, these models demonstrate remarkable capabilities in capturing generalizable mobility laws across diverse urban environments~\cite{zhou2024urban}. 

Formally, given a trajectory sequence $\mathcal{S}$, the MFM learns the joint probability distribution by decomposing it into conditional probabilities:
\begin{equation}
    P_{\theta}(\mathcal{S}) = \prod_{i=1}^{n} P_{\theta}(l_i \mid l_{<i}, t_{\le i})
\end{equation}
where $l_{<i}$ denotes the history of visited locations. The objective is to maximize the likelihood of observed trajectories across distinct regions, enabling the model to achieve strong zero-shot generalization capabilities in unseen cities. Despite their robust statistical modeling, traditional MFMs primarily operate on numerical coordinates, motivating the need to bridge their latent representations with semantic reasoning.

\subsection{Task Definitions}

\paragraph{Task 1: Mobility Prediction (Next-Location).}
Given a user's historical trajectory $\mathcal{S}_{hist} = \{(l_1, t_1), \dots, (l_k, t_k)\}$, the goal is to predict the exact next location $l_{k+1}$ at time $t_{k+1}$. This task evaluates the model's ability to capture sequential dependencies and user inertial habits:
\begin{equation}
    \hat{l}_{k+1} = \arg\max_{l \in \mathcal{L}} P(l \mid \mathcal{S}_{hist})
\end{equation}

\paragraph{Task 2: Mobility Understanding.}
This task requires the model to bridge the gap between numerical coordinates and semantic intent. Given a continuous coordinate sequence $\mathcal{S}$, the model must generate natural language text $Y_{desc}$ that accurately summarizes the user's high-level spatio-temporal behavioral patterns:
\begin{equation}
    \hat{Y}_{desc} = \arg\max_{Y} P(Y \mid \mathcal{S})
\end{equation}

\paragraph{Task 3: Standard Trajectory Generation.}
Unlike single-step prediction, this task aims to synthesize an entire continuous sequence of future movements. Given a user's historical trajectory $\mathcal{S}_{hist}$, the goal is to generate a physically plausible future trajectory $\mathcal{S}_{future} = \{(l_{k+1}, t_{k+1}), \dots, (l_{k+m}, t_{k+m})\}$ that adheres to the user's inherent behavioral patterns:
\begin{equation}
    \hat{\mathcal{S}}_{future} = \arg\max_{\mathcal{S}} P(\mathcal{S} \mid \mathcal{S}_{hist})
\end{equation}
This standard generation task relies primarily on historical inertia to extrapolate future states. However, it faces limitations when environmental conditions change unexpectedly.

\paragraph{Task 4: Counterfactual Trajectory Generation.}
Building upon standard generation, this task evaluates the model's ``what-if'' reasoning capabilities. We frame this fundamentally as conditional trajectory generation under novel constraints, rather than strictly mathematical causal inference. Given a user's historical trajectory $\mathcal{S}_{hist}$ and a set of newly introduced, unseen spatiotemporal constraints $\mathcal{C}$ (such as a necessary detour due to a sudden traffic accident), the model must synthesize a modified trajectory $\mathcal{S}_{future}$:
\begin{equation}
    \hat{\mathcal{S}}_{future} = \arg\max_{\mathcal{S}} P(\mathcal{S} \mid \mathcal{S}_{hist}, \mathcal{C})
\end{equation}
This requires the model to break away from established distributions, satisfying the novel external constraints while still preserving realistic human mobility laws. When $\mathcal{C}$ is empty, this task gracefully degrades to standard trajectory generation.

\section{Methodology}

To bridge the fundamental gap between the statistical fidelity of Mobility Foundation Models (MFMs) and the semantic reasoning of Large Language Models (LLMs), we propose \textbf{MoveFM-R}. As illustrated in Figure~\ref{fig:framework}, the framework operates by taking a trajectory sequence and a natural language instruction corresponding to a specific task as inputs. The trajectory sequence is transformed via the universal codebook and processed by the MFM to produce spatiotemporal representations. These features are then mapped through a projection layer and fed into the LLM alongside the text prompt, enabling the model to generate semantically and geographically coherent responses.
The two core stages of MoveFM-R are
(1) Bridging the comprehension gap: We align MFM latent spatial features with the LLM’s semantic space using a universal multi-layer codebook and a progressive alignment curriculum.
(2) Resolving the ``what-if'' generation dilemma: We reframe counterfactual generation as a trajectory editing task, utilizing a self-reflective reinforcement learning mechanism to satisfy novel constraints while preserving mobility realism.

\subsection{Bridging the Comprehension Gap}
\subsubsection{Universal Codebook Construction}
To effectively integrate MFMs and LLMs, we must first resolve the vocabulary mismatch between continuous coordinates and discrete language tokens. 
Traditional methods that map coordinates to unique IDs (from $0$ to $N$) face a critical bottleneck: the vastness of geographic space leads to a vocabulary explosion. 
Critically, the resulting severe data sparsity creates a fragmented embedding space where the LLM struggles to learn meaningful representations for the long tail of low-frequency locations due to the power-law distribution of visit frequencies and the lack of shared semantic context in discrete IDs.

To overcome these limitations, we propose a Universal Multi-layer Codebook. 
By decomposing locations into hierarchical combinations of discrete codes based on their semantics and surrounding environmental information, we compress the vast coordinate space into a compact vocabulary.
Crucially, this design enhances representation learning for rare locations via knowledge transfer: since a low-frequency location shares constituent codewords with high-frequency locations at various layers, the model can infer its properties from these shared semantic patterns.
Furthermore, our codebook supports robust zero-shot generalization. By training on multi-city data, the codebook captures a city-agnostic semantic syntax (e.g., recognizing that a code combination represents a commercial district regardless of cities), enabling the model to encode unseen locations based on learned structural similarities.

\textbf{Hybrid Semantic-Spatial Encoding.}
We employ a Hybrid Semantic-Spatial Encoding strategy to construct the input representation. 
Different from previous codebook construction approaches (e.g., QT-Mob~\cite{chen2025enhancing}) that rely solely on textual semantics for quantization, we identify a critical limitation.
In dense urban environments, spatially adjacent but distinct locations (e.g., two different shops in the same street) likely share nearly identical address descriptions and POI contexts. 
Consequently, relying purely on encoded text vectors leads to codebook conflicts, where distinct physical locations collapse into identical codes.
To resolve this, we construct a distinguishable input representation $E$ by fusing two modalities:
\begin{enumerate}%
    \item Textual Semantic Vector ($v_{\text{text}}$): We encode descriptive attributes (e.g., addresses) via a pre-trained encoder~\cite{zhang2025qwen3}. Note that these features are extracted offline to ensure training efficiency.
    \item Geometric Position Vector ($v_{\text{geo}}$): We explicitly encode coordinates using high-dimensional sinusoidal embeddings. These embeddings provide a position-sensitive inductive bias, ensuring that even if two locations share indistinguishable textual semantics, their fused representations $E = \text{Concat}(v_{\text{text}}, v_{\text{geo}})$ remain separable in the high-dimensional space.
\end{enumerate}

\textbf{Hierarchical Discretization via RQ-VAE.}
The fused input $E$ is then discretized using a Residual Quantized Variational Autoencoder (RQ-VAE)~\cite{lee2022autoregressive}. 
Specifically, the quantization process operates recursively: for an input vector $E$, the model performs a greedy search at the first layer to find the nearest codeword, subtracts it to obtain a residual, and recursively quantizes this residual at subsequent layers.
Formally, at depth $n$, given the residual $r_n$ (where $r_0 = E$), the discrete code index $c_n$ is selected as:
\begin{equation}
    c_n = \arg\min_{k} \| r_n - v^n_k \|_2^2, \quad \text{and} \quad r_{n+1} = r_n - v^n_{c_n},
\end{equation}
where $\mathcal{C}^n = \{v^n_k\}_{k=1}^{K}$ denotes the codebook at layer $n$. This hierarchical structure produces a tuple of tokens $\{c_1, \dots, c_N\}$.

\textbf{Spatially-Aware Optimization.}
To optimize the codebook, we employ a composite objective comprising three parts.
First, we adopt the standard objectives inherited from the RQ-VAE framework to ensure faithful codebook alignment:

\noindent \textbf{Feature Reconstruction Loss ($\mathcal{L}_{\text{rec}}$).} 
We minimize the distance between the hybrid input $E$ and the reconstructed vector $\hat{E}$ (defined as the sum of selected code vectors $\sum_{n=1}^{N} v^n_{c_n}$) to preserve fused semantic-spatial information:
\begin{equation}
    \mathcal{L}_{\text{rec}} = \| E - \hat{E} \|_2^2.
\end{equation}

\noindent \textbf{Quantization Loss ($\mathcal{L}_{\text{RQ}}$).} 
To update the codebook, we employ the standard quantization loss, which pulls the codebook embeddings towards the encoder residuals and vice versa using the stop-gradient operator ($\text{sg}[\cdot]$):
\begin{equation}
    \mathcal{L}_{\text{RQ}} = \sum_{n=1}^{N} \left( \| \text{sg}[r_n] - v^n_{c_n} \|_2^2 + \beta \| r_n - \text{sg}[v^n_{c_n}] \|_2^2 \right).
\end{equation}

\noindent \textbf{Geometric Reconstruction Loss ($\mathcal{L}_{\text{geo}}$).} 
To explicitly enforce spatial grounding, we introduce an auxiliary regression task.
A projection head predicts the normalized coordinates $(\hat{\text{lat}}, \hat{\text{lon}})$ from the quantized feature $\hat{E}$. We optimize this using the Smooth-L1 loss~\cite{girshick2015fast}:
\begin{equation}
    \mathcal{L}_{\text{geo}} = \text{SmoothL1}(\text{MLP}_{\text{geo}}(\hat{E}), \text{Coord}_{\text{gt}}).
\end{equation}
This loss compels the discrete code combinations to retain precise coordinate information, preventing the quantization process from abstracting away essential high-frequency spatial details.
The final training objective is $\mathcal{L} = \mathcal{L}_{\text{rec}} + \mathcal{L}_{\text{RQ}} + \gamma \mathcal{L}_{\text{geo}}$, where \( \gamma \) is set to 0.2 in practice.

\subsubsection{Progressive Alignment Curriculum: From Token to Trajectory}
\label{sec:curriculum}

Although the Universal Codebook unifies the discrete vocabulary, a fundamental \textbf{representation gap} persists between the MFM's intrinsic representation space and the LLM's semantic space. To bridge this, we propose a \textbf{Progressive Alignment Curriculum}, as illustrated in Figure~\ref{fig:framework}, systematically guiding the LLM to interpret MFM features.

\textbf{Location Semantic Understanding.} We begin by establishing a bidirectional semantic understanding of \textbf{Code Tokens}. The model is trained to mutually map between discrete token of locations and their geographic descriptions. This interaction effectively grounds the abstract MFM tokens in the LLM's semantic knowledge, creating a robust semantic anchor for subsequent sequence processing.

\textbf{Trajectory Semantic Understanding.} Moving to the trajectory level, we employ a lightweight MLP adapter to project the MFM's latent vectors into the LLM's embedding dimension. To ensure substantive content alignment, we design a task curriculum with increasing difficulty:

\begin{enumerate}%
    \item Raw Trajectory Reconstruction: Given the projected MFM sequence features, the LLM is tasked with translating them back into a factual textual description of the visit sequence. This enforces the precise retention of low-level movement details.
    \item Spatiotemporal Feature Summarization: The LLM reasons over the same feature input to infer abstract high-level patterns, such as identifying users' preferred locations and modeling the temporal evolution of movement probabilities.
\end{enumerate}

\textbf{Understanding as a Prerequisite.} Crucially, this high-level trajectory understanding is integrated as a \textbf{prerequisite} for both downstream prediction and generation tasks in practice. We formulate prediction and generation as a conditional multi-step objective: the LLM is mandated to output a spatiotemporal trajectory feature summary \textit{before} providing the final result. Analogous to Chain-of-Thought (CoT) prompting~\cite{wei2022chain}, this mechanism establishes a coherent ``Understanding $\rightarrow$ Prediction $|$ Generation'' chain, ensuring that outputs are grounded in inferred mobility features rather than shallow sequence matching. Detailed prompt designs are provided in the supplementary materials. The detailed prompt templates used in our framework are provided in Section S1 of the Supplementary Material.

\textbf{Unified Training Objective.} Finally, we unify the tasks of understanding, prediction, and generation under a single training framework. Formally, let $X_{seq}$ denote the input trajectory sequence and $g_{\phi}$ be the MFM encoder. The aligned representation is obtained via $H_{seq} = \text{MLP}(g_{\phi}(X_{seq}))$. Note that because the MFM and LLM share the same underlying codebook tokens, the feature spaces are inherently aligned, allowing a simple MLP to bridge the remaining modality gap without complex alignment modules effectively. We standardize all tasks into a unified input-output stream using task-specific instruction prompts $I_{task}$. The model is optimized to maximize the likelihood of the target sequence $Y$ (encompassing both the analysis summary and the final result) using the standard next-token-prediction cross-entropy loss:
\begin{equation}
\mathcal{L} = -\frac{1}{N} \sum_{t=1}^{N} \log P_{\theta}(y_t | y_{<t}, I_{task}, H_{seq})
\end{equation}
This unified objective allows the model to seamlessly transfer the comprehension and analysis capabilities acquired during the alignment phase to downstream mobility tasks.

\subsection{Resolving the Generation Dilemma via Trajectory Editing}

While the Progressive Alignment Curriculum bridges the comprehension gap, enabling the model's trajectory prediction and generation capabilities, a critical challenge persists: the ``What-if'' Generation Dilemma, where models struggle to maintain a dynamic trade-off between fidelity to historical habits and compliance with novel spatiotemporal constraints. 

Inspired by image editing tasks in computer vision—where a model modifies an image to match a prompt while preserving its original structure—we reframe counterfactual mobility generation as a \textbf{Trajectory Editing} task. Instead of generating a trajectory from scratch, the model learns to ``edit'' a baseline trajectory, initially generated solely based on historical mobility patterns, to satisfy new environmental constraints. 

A major hurdle in training this editing mechanism is the lack of real-world ``counterfactual'' data. To circumvent this, we design a self-supervised strategy to construct a dataset that pairs historical habits with new constraints. Our key insight is that the spatiotemporal feature discrepancy between a user's past behavior and their actual future behavior can be viewed as an implicit ``environmental instruction'' ($\Delta F$). Formally, given a user's history ($T_{hist}$) and the ground-truth future trajectory ($T_{target}$), we formulate the training instance as follows:
\begin{itemize}
    \item \textbf{History Features ($F_{hist}$):} We extract spatiotemporal features from $T_{hist}$, representing the user's established routine.
    \item \textbf{Target Features ($F_{target}$):} We extract features from $T_{target}$, representing the actual behavior under specific conditions.
    \item \textbf{Editing Instruction ($\Delta F$):} We calculate the difference $\Delta F = F_{target} - F_{hist}$, which serves as the editing instruction. This requires the model to adjust the baseline trajectory to match the specific properties of the target day ($F_{target}$).
\end{itemize}

\begin{figure}[h]
    \centering
    \includegraphics[width=0.85\linewidth]{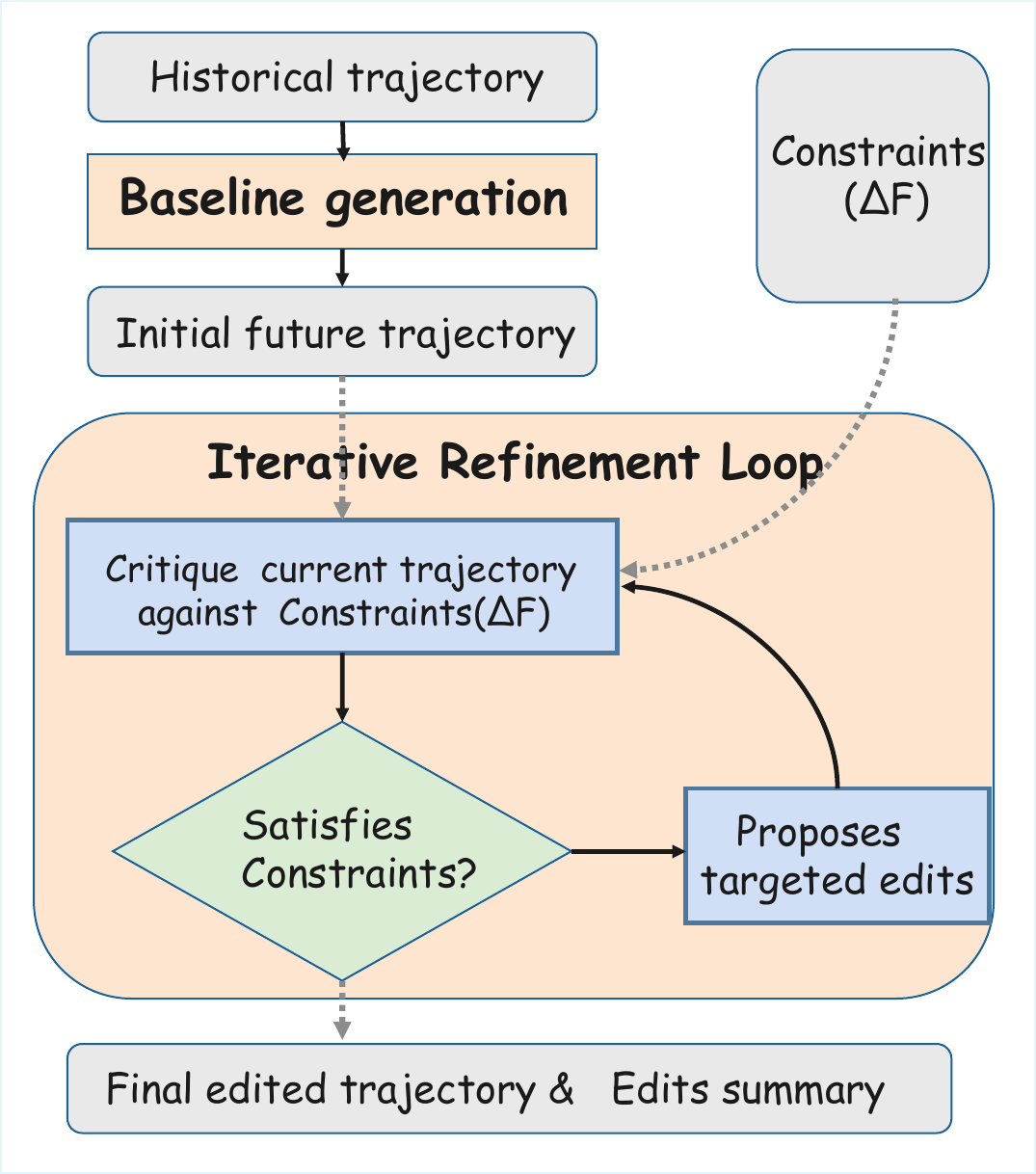}
    \vspace{-2mm}
    \caption{Self-reflection and reasoning process}
    \label{fig:Self-reflection}
\end{figure} 

\subsubsection{Self-Reflective Reasoning Mechanism}
Our \textbf{Self-Reflective Reasoning} operationalizes the ``minimal edits'' principle through a structured, iterative process to accomplish the trajectory editing task, as illustrated in Figure~\ref{fig:Self-reflection}.
\begin{enumerate}
    \item \textbf{Baseline Generation:} The model first generates an initial trajectory based solely on historical data. This serves as a ``zero-scenario'' baseline, reflecting the user's inertial behavior without external interference.
    \item \textbf{Iterative Refinement Loop:} The model then enters a deterministic refinement loop. At each step, it critiques the current trajectory against the constraints defined by $\Delta F$. If statistical mismatches are detected, the model proposes targeted edits within a discrete-continuous hybrid action space $\mathcal{A}$. Specifically, a single editing operation consists of one of the following:
    \begin{itemize}
        \item \textbf{Insertion:} Augmenting the current sequence by adding a trajectory point at a specific index to increase granularity.
        \item \textbf{Pruning:} Removing an existing point from the trajectory to eliminate redundant or noisy information.
        \item \textbf{Refinement:} Modifying the spatio-temporal coordinates (i.e., time and location) of a pre-existing point to optimize the path.
    \end{itemize}
    \item \textbf{Termination with Minimal Edits:} The loop terminates once the trajectory satisfies all constraints with minimal edits or reaches the step limit.
\end{enumerate}
By explicitly optimizing for the fewest number of edits, we ensure the final output is a synthesis that respects the user's history rather than a hallucinated trajectory.

\subsubsection{Optimizing Self-Reflective via GRPO}
To effectively train the model to execute the iterative reasoning loop defined above, we use Group Relative Policy Optimization (GRPO)~\cite{shao2024deepseekmath} to implement trajectory reasoning.
Unlike traditional algorithms like PPO~\cite{schulman2017proximal} that require a separately trained value function, GRPO computes advantages using group-relative rewards. This significantly reduces memory and computational overhead, making it particularly suitable for training LLMs. 
The model is trained to perform iterative optimization editing following the structured reasoning template detailed in table~\ref{tab:temp_promptl}.
\begin{table}[h]
    \centering
    \caption{Template for Self-Reflective Reasoning with GRPO}
    \begin{tabular}{p{0.95\linewidth}}
        \hline
        Please answer the following questions step by step.\ 
        You need to think and reason before answering, outputting your reasoning process between \texttt{<think>} and \texttt{</think>}, and providing your final answer between \texttt{<answer>} and \texttt{</answer>}. \
        
        \textcolor{cyan}{Input}: Historical trajectory data, initial generated trajectory, spatiotemporal constraints. \

        \textcolor{brown}{Task}: Modify the initial trajectory data based on the historical data and the spatiotemporal constraints of the scene. Ensure that the modified trajectory conforms to the given statistical spatiotemporal characteristics and uses the minimum modification step size. \\
        \hline
    \end{tabular}
    \label{tab:temp_promptl}
\end{table}

The design of the reward function is guided by the objective of \textbf{distributional consistency}. Unlike common reasoning tasks that target an Exact-Match (EM) solution, mobility generation is stochastic. Our goal is to generate trajectories that align with the correct statistical distribution. Consequently, we formulate a reward function based on matching key spatiotemporal properties (e.g., visit probabilities by time period and location) rather than coordinate-level matching.
Let $\phi(\tau)$ denote the statistical feature vector of trajectory $\tau$. The distribution reward is:
\begin{equation}
R_{\text{distribution}}(\tau) = \sum_{k=1}^K \mathbf{1}\!\left[\phi_k(\tau) = \phi_k(\tau^{*})\right],
\end{equation}
where $\tau^{*}$ is the ground truth trajectory. Each matched feature contributes $+1$. Additionally, to avoid unrealistic length deviations, we penalize discrepancies between generated and ground-truth lengths:
\begin{equation}
R_{\text{length}}(\tau) = - \frac{\left|\,|\tau| - |\tau^{*}|\,\right|}{|\tau^{*}|}.
\end{equation}
In summary, the total reward can be expressed as follows: $R(\tau) = R_{\text{distribution}}(\tau) + R_{\text{length}}(\tau).$
To further stabilize the training process, we employ Supervised Fine-Tuning (SFT) as a warm-up phase prior to GRPO.

\begin{table*}[t]
    \caption{Experiment Result on Next Location Prediction Task (HR@1).}
    \label{tab:city-overall}
    
    \centering
    
    \footnotesize
    
    \begin{tabular*}{0.85\textwidth}{@{\extracolsep{\fill}} l *{9}{c} }
        \toprule
        & DeepMove & GETNext & TrajFM & Unitraj & TrajMoE & Mobility-LLM & QT-Mob & \textbf{MoveFM-R} & Improve \\
        \midrule
        Atlanta     & 0.171 & 0.178 & 0.196 & 0.210 & \underline{0.245} & 0.214 & 0.240 & \textbf{0.281} & +14.7\% \\
        Chicago     & 0.188 & 0.189 & 0.212 & 0.219 & 0.269             & 0.218 & \underline{0.306} & \textbf{0.334} & +9.2\% \\
        Seattle     & 0.220 & 0.227 & 0.255 & 0.283 & 0.309             & 0.270 & \underline{0.315} & \textbf{0.368} & +16.8\% \\
        Washington  & 0.204 & 0.197 & 0.202 & 0.215 & 0.265             & 0.224 & \underline{0.286} & \textbf{0.328} & +14.7\% \\
        \bottomrule
    \end{tabular*}
    
\end{table*}

\begin{table*}[t]
    \centering
    \caption{Experiment result on zero-shot and few-shot settings of next location prediction (HR@1).}
    \label{tab:city-zeroshot}
    
    \footnotesize
    
    \begin{tabular*}{0.85\textwidth}{@{\extracolsep{\fill}} l *{8}{c}}
        \toprule
        \multirow{2}{*}{Method} & 
        \multicolumn{2}{c}{Atlanta} & 
        \multicolumn{2}{c}{Chicago} & 
        \multicolumn{2}{c}{Seattle} & 
        \multicolumn{2}{c}{Washington} \\
        
        \cmidrule(lr){2-3} \cmidrule(lr){4-5} \cmidrule(lr){6-7} \cmidrule(lr){8-9} 
        
        & Zero-shot & Few-shot & Zero-shot & Few-shot & Zero-shot & Few-shot & Zero-shot & Few-shot \\
        \midrule
        
        TrajMoE & 0.121 & 0.151 & 0.085 & 0.098 & 0.146 & 0.194 & 0.141 & 0.168 \\
        QT-Mob  & \underline{0.132} & \underline{0.203} & \underline{0.242} & \underline{0.255} & \underline{0.218} & \underline{0.244} & \underline{0.242} & \underline{0.271} \\
        Ours    & \textbf{0.164} & \textbf{0.264} & \textbf{0.280} & \textbf{0.309} & \textbf{0.262} & \textbf{0.294} & \textbf{0.272} & \textbf{0.292} \\
        \midrule
        
        Improve & +24.24\% & +30.05\% & +15.70\% & +21.18\% & +20.18\% & +20.49\% & +12.40\% & +7.75\% \\
        \bottomrule
    \end{tabular*}
\end{table*}

\section{Experiments} 

\subsection{Experimental Setup}

\subsubsection{Dataset}
We evaluate our method on four real-world human mobility datasets spanning Atlanta, Chicago, Seattle, and Washington, D.C. The statistical overview of the datasets used is presented in Table~\ref{tab:dataset_stats}. The spatial domain of each city is discretized into $500\text{m} \times 500\text{m}$ grid cells with a temporal resolution of 30 minutes. User trajectories are constructed using a three-day sliding window. To mitigate sparsity and noise, we filter out trajectories containing fewer than five trips, while sequences exceeding 145 points are truncated to retain the most recent 145. Each visited location is augmented with rich semantic context (e.g., Points of Interest) derived from OpenStreetMap (OSM). We strictly adhere to ethical guidelines regarding user privacy and have anonymized the datasets.

\textbf{Dataset Partitioning.} For non-LLM baselines, we partition users into training, validation, and testing sets using a 60\%, 20\%, 20\% split. However, due to limited computational resources, all LLM-based methods are trained using only the first 40,000 samples from the standard training set. Crucially, we maintain the same testing set across all methods to ensure a fair and consistent evaluation of final performance metrics.

\begin{table}[h!]
    \centering
    \caption{Statistical information for the trajectory datasets used in our experiments.}
    \label{tab:dataset_stats}
    \begin{tabular}{l c r r}
        \toprule
        \textbf{City} & \textbf{Duration} & \textbf{Locations} & \textbf{Trajectories} \\
        \midrule
        Atlanta & 7 days & 1,175 & 200,000 \\
        Chicago & 7 days & 4,166 & 200,000 \\
        Seattle & 7 days & 1,046 & 200,000 \\
        Washington & 7 days & 1,361 & 200,000 \\
        \bottomrule
    \end{tabular}
\end{table}

\subsubsection{Evaluation Metrics.} For prediction task, we adopt the widely used metric Hit Rate (HR@$1$) to evaluate prediction accuracy~\cite{han2025trajmoe,chen2025enhancing}. 
For generation task, we adopt commonly used metrics $BLEU$, $TVD$, and $JSD$ to measure the time and location similarity between the generated sequence and the real sequence respectively~\cite{reed2016generative,wang2024large}. For more details on all metrics above, 
please refer to Section S2 in the Supplementary Material.

\subsubsection{Baselines.} Our baseline selection spans different methodological families to ensure a comprehensive evaluation. 

For the \textbf{prediction task}, we selected the following approaches:
\begin{itemize}
    \item \textbf{DeepMove}~\cite{feng2018deepmove} is an attentional recurrent neural network that captures both long-term periodic patterns and short-term sequential regularities in user mobility.
    \item \textbf{TrajBert}~\cite{si2023trajbert} adapts the powerful BERT architecture to model trajectories by treating locations as tokens and learning deep, bidirectional contextual representations for prediction.
    \item \textbf{GETNext}~\cite{yang2022getnext} integrates a graph neural network to explicitly learn spatial relationships between locations with a Transformer-based encoder to capture complex spatio-temporal dependencies.
    \item \textbf{TrajFM}~\cite{lin2024trajfm} is a foundation model for trajectories that is pre-trained on a massive dataset to learn universal mobility patterns adaptable to various downstream tasks.
    \item \textbf{Unitraj}~\cite{zhu2024unitraj} is a universal pre-trained model that unifies the representation of diverse trajectory data types, including spatio-temporal points, semantic texts, and graph structures.
    \item \textbf{TrajMoE}~\cite{han2025trajmoe} employs a Mixture-of-Experts (MoE) architecture where different ``expert'' sub-networks specialize in modeling distinct mobility patterns for more accurate and robust predictions.
    \item \textbf{Mobility-LLM}~\cite{gong2024mobility} is a large language model-based framework that reformulates trajectory prediction as a language modeling task by converting mobility data into textual sequences.
    \item \textbf{QT-Mob}~\cite{chen2025enhancing} enhances LLMs for mobility prediction by incorporating a query-time adaptation mechanism that retrieves and integrates relevant external spatio-temporal knowledge at the time of inference.
\end{itemize}

For the \textbf{generation task}, we selected the following approaches:
\begin{itemize}
    \item \textbf{DiffTraj}~\cite{zhu2023difftraj} applies a denoising diffusion probabilistic model to generate realistic and diverse human trajectories by progressively refining a random noise signal into a structured sequence.
    \item \textbf{Marionette}~\cite{deng2025marionette} is a controllable trajectory generation model based on guided diffusion, allowing for the synthesis of trajectories that adhere to specific user-defined constraints or conditions.
    \item \textbf{COPB}~\cite{shao2024chain} leverages the Chain-of-Thought prompting technique with large language models to iteratively reason about user preferences and construct plausible, context-aware trajectories.
    \item \textbf{LLMob}~\cite{wang2024large} is a comprehensive framework that utilizes the generative and reasoning capabilities of large language models to produce human-like trajectories based on user profiles and historical data.
\end{itemize}

\subsubsection{Implementation Details.} Experiments were primarily conducted on a cluster of four NVIDIA A800 (40GB) GPUs. We employed Qwen2.5-7B~\cite{hui2024qwen2} as the backbone LLM and TrajMOE~\cite{han2025trajmoe} as the mobility foundation encoder. To optimize computational efficiency, we utilized LoRA fine-tuning~\cite{hu2022lora} and parallel training acceleration. The models were optimized using the AdamW optimizer with a cosine annealing learning rate scheduler. Specifically, the maximum learning rate was set to 1e-4, while the initial and minimum warmup learning rates were 2e-5, with a warmup ratio of 0.03. All experiments were conducted for a maximum of 5 epochs with a total batch size of 96, and the best-performing model on the validation set was selected for testing. To mitigate the variance introduced by random initialization, we repeated each experiment 5 times with different random seeds and report the average performance as the final result. For the self-reflective reasoning stage (GRPO), we employed an auxiliary setup using two NVIDIA A100 (80GB) GPUs with Qwen3-4B~\cite{yang2025qwen3} serving as the backbone LLM to facilitate efficient reinforcement learning. For information on the specific model parameters, 
please refer to Section S3 in the Supplementary Material.

\subsection{Mobility Prediction}
\textbf{Next Location Prediction.} 
We evaluated the performance of all methods on four benchmark datasets. 
To fully leverage the benefits of pre-training, methods supporting cross-city learning (e.g., TrajMoE) were trained on a mixed dataset comprising all four cities before being tested on each specific city. 
The results, summarized in Table~\ref{tab:city-overall}, yield three key observations:

\begin{enumerate}%
    \item \textbf{Superior Overall Performance:} MoveFM-R consistently outperforms all baselines, achieving an average accuracy improvement of over 10\% across all datasets.
    
    \item \textbf{Enhancing the MFM Backbone:} Compared to TrajMoE (which serves as our mobility encoder), MoveFM-R achieves a substantial improvement of approximately 20.4\%. 
    This significant margin validates our core hypothesis: the statistical power of MFMs is constrained by a lack of semantic understanding, and our framework successfully unlocks this potential by integrating LLM-driven reasoning.
    
    \item \textbf{Necessity of Spatiotemporal Features:} Our approach outperforms pure LLM-based baselines (which rely solely on plain text input) by an additional 13.9\%. 
    This highlights the limitation of treating mobility purely as a language task and emphasizes the critical value of grounding LLMs in domain-specific spatiotemporal representations captured by the MFM.
\end{enumerate}

\textbf{Zero-Shot and Few-Shot Performance.}
To evaluate the model's generalization capability across geographic domains, we conducted zero-shot experiments (training on three cities and testing on the fourth novel city) and few-shot experiments (fine-tuning on only 500 examples of the target city). The results, presented in Table~\ref{tab:city-zeroshot}, reveal several critical insights:

First, our approach consistently outperforms both the strongest sequence-based (TrajMoE) and LLM-based (QT-Mob) baselines in terms of zero-shot and few-shot performance across all four cities. This robust generalization stems from the synergy between our \textbf{Universal Semantic Codebook} and the \textbf{Mobility Understanding Curriculum}. The universal codebook allows the model to map unseen locations to familiar semantic tokens, while the understanding curriculum enables the effective transfer of underlying spatiotemporal mobility patterns. Together, these components ensure that both the geographic semantics (the "where") and the movement dynamics (the "how") are successfully generalized to novel environments.
Second, the LLM-based approach, QT-Mob, comprehensively outperforms the purely sequence-based model, TrajMoE. This highlights the inherent ability of language models to transfer semantic knowledge across diverse urban environments. Notably, our approach achieves zero-shot accuracy in three cities (Chicago,Seattle,Washington) that surpasses the classic method, DeepMove, even when the latter is fine-tuned on the full dataset. This further emphasizes the strong generalization capabilities of our method, demonstrating that synthesizing statistical fidelity with semantic reasoning captures universal mobility laws that transcend specific city boundaries.

\subsection{Mobility Generation }

\textbf{Trajectory Generation.}
For generation tasks, we prioritize distributional fidelity between generated and real sequences over point-matching accuracy. We evaluated all methods on four city datasets, where the task involves generating a user's third-day trajectory conditioned on historical data from the previous two days. The results, detailed in Table~\ref{tab:overall-gene}, yield several key observations.

First, MoveFM-R achieves state-of-the-art performance across all metrics (BLEU, TVD, and JSD) for both temporal and location distributions, demonstrating its effectiveness in modeling spatiotemporal mobility patterns. Second, by leveraging the MFM to extract informative features from numerical sequences, our method significantly outperforms LLM-based baselines (COPB, LLMob). This highlights the importance of grounding LLM reasoning in domain-specific representations rather than pure text. Furthermore, our approach surpasses pure sequence modeling methods (DiffTraj, Marionette) by benefiting from the semantic understanding and reasoning capabilities of LLMs. Collectively, these results demonstrate that integrating structured trajectory features with LLMs offers consistent advantages over both traditional architectures and LLM-only methods.

\begin{figure}[ht]
    \centering
    \includegraphics[width=\linewidth]{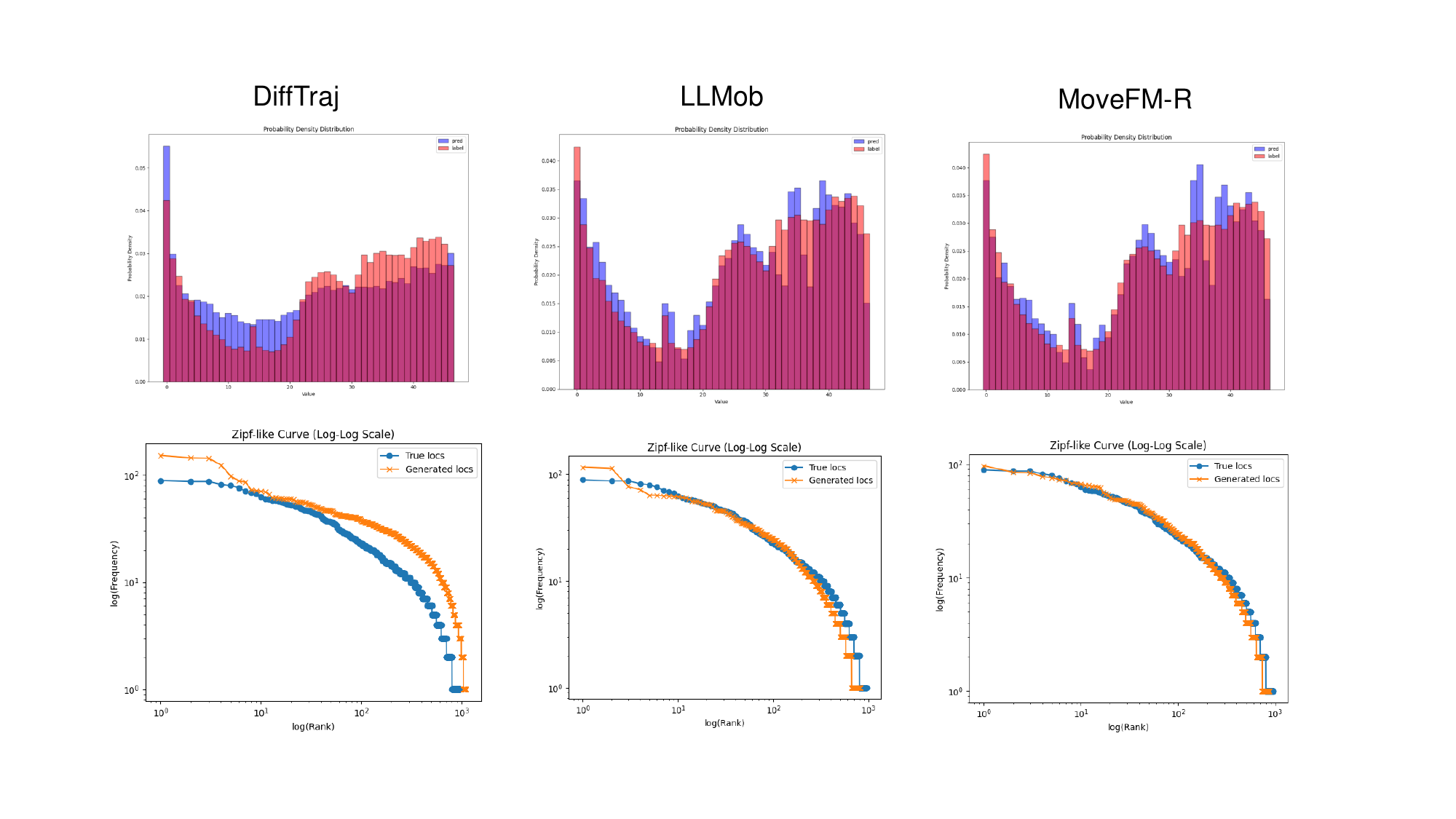} %
    \vspace{-10mm}
    \caption{
        \textbf{Comparison of Temporal and Location Distributions.} 
        We evaluate the distributions of generated trajectories from our model (MoveFM-R) against baselines (DiffTraj, LLMob). 
        \textbf{Top row:} Visualization of the temporal distribution. The generated distribution (red) from our model more accurately matches the true temporal distribution (blue) of user activities over time. 
        \textbf{Bottom row:} Visualization of the location distribution on a log-log scale (Zipf-like plot). The curve for our generated data (orange) shows a much tighter fit to the ground-truth data (blue) across the entire spectrum, from popular (head) to rare (tail) locations.
    }
    \label{fig:dist_visualization}
\end{figure}

To qualitatively validate this distributional fidelity, we visualize the temporal and location distributions of trajectories generated by representative algorithms (Figure~\ref{fig:dist_visualization}). Compared to the baselines, the trajectories generated by MoveFM-R align much more closely with the ground-truth distributions. Notably, for the location distribution, our method exhibits significant improvements across both high-frequency (head) and long-tail regions. This visual evidence strongly corroborates our quantitative results, confirming that our framework achieves superior fidelity to real-world mobility patterns—a critical prerequisite for robust conditional trajectory generation.

\begin{table}[h] 
    \centering
    \caption{\normalfont Performance of unconditional trajectory generation task.}
    \label{tab:overall-gene}
    \resizebox{\linewidth}{!}{%
        \begin{tabular}{l *{2}{ccc}}
            \toprule
            \multirow{2}{*}{\textbf{Method}} &
            \multicolumn{3}{c}{\textbf{Time}} &
            \multicolumn{3}{c}{\textbf{Location}} \\
            \cmidrule(lr){2-4} \cmidrule(lr){5-7}
            & \textbf{$\text{Bleu} \uparrow$} & \textbf{$\text{TVD} \downarrow$} & \textbf{$\text{JSD} \downarrow$} & \textbf{$\text{Bleu} \uparrow$} & \textbf{$\text{TVD}\downarrow$}  & \textbf{$\text{JSD} \downarrow$} \\
            \midrule
            DiffTraj      & 0.387 & 0.117 & 0.009 & 0.076 & 0.494 & 0.220 \\
            Marionette    & 0.582 & 0.082 & 0.008 & 0.092 & 0.346 & 0.102 \\
            COPB          & 0.426 & 0.096 & 0.009 & 0.084 & 0.382 & 0.133 \\
            LLMob         & 0.605 & 0.085 & 0.007 & 0.095 & 0.323 & 0.095 \\
            \textbf{Ours} & \textbf{0.628} & \textbf{0.064} & \textbf{0.006} & \textbf{0.136} & \textbf{0.250} & \textbf{0.062} \\
            \bottomrule
        \end{tabular}%
    }
\end{table}

\subsection{Ablation Study}

\begin{table}[t]
    \centering
    \caption{Ablation study on next location prediction task.(HR@1)}
    \label{tab:Ablation-pre}
    \footnotesize
    \setlength{\tabcolsep}{0pt}
    
    \begin{tabular*}{\linewidth}{@{\extracolsep{\fill}} l cccc }
        \toprule
        Method & Atlanta & Chicago & Seattle & Washington \\
        \midrule
        \textbf{Ours} & \textbf{0.281} & \textbf{0.334} & \textbf{0.368} & \textbf{0.328} \\
        
        w/o CB  & 0.243   & 0.310   & 0.326   & 0.306 \\
        w/o RU  & 0.270   & 0.328   & 0.350   & 0.314 \\
        w/o FM  & 0.259   & 0.318   & 0.337   & 0.304 \\
        \bottomrule
    \end{tabular*}
\end{table}

\begin{table}[t]
    \centering
    \caption{Ablation on unconditional trajectory generation task.}
    \label{tab:Ablation-gene}
    
    \footnotesize
    
    \setlength{\tabcolsep}{0pt}
    
    \begin{tabular*}{\linewidth}{@{\extracolsep{\fill}} l cccccc }
        \toprule
        \multirow{2}{*}{Method} & 
        \multicolumn{3}{c}{Time} & 
        \multicolumn{3}{c}{Location} \\
        
        \cmidrule(lr){2-4} \cmidrule(lr){5-7} 
        
        & Bleu $\uparrow$ & TVD $\downarrow$ & JSD $\downarrow$ & Bleu $\uparrow$ & TVD $\downarrow$ & JSD $\downarrow$ \\
        \midrule
        
        \textbf{Ours} & \textbf{0.628} & \textbf{0.064} & \textbf{0.006} & \textbf{0.136} & \textbf{0.250} & \textbf{0.062} \\
        
        w/o CB  & 0.598 & 0.090 & 0.007 & 0.112 & 0.273 & 0.072 \\
        w/o RU  & 0.613 & 0.072 & 0.006 & 0.108 & 0.265 & 0.068 \\
        w/o FM  & 0.594 & 0.087 & 0.007 & 0.108 & 0.278 & 0.074 \\
        \bottomrule
    \end{tabular*}
    
\end{table}

To validate the effectiveness of each component in our framework, we conducted ablation studies on four datasets. We evaluated the model under four settings: (i) without CB (codebook), (ii) without RU (representation understanding), (iii) without FM (base model).
The results for the prediction task are in Table~\ref{tab:Ablation-pre}, and for the generation task in Table~\ref{tab:Ablation-gene}. The following are the key observations:

First, removing both the base model and the codebook results in a significant drop in performance, highlighting the importance of the spatiotemporal trajectory features and spatial semantics provided by the base model and the structured position encoding. Second, removing representation understanding results in a moderately consistent drop in performance on both tasks, highlighting that fine-grained feature understanding helps the LLM better exploit spatiotemporal information. This effect is slightly more pronounced in the generation task.
Overall, these ablation results confirm that each component makes a meaningful contribution and that they collectively enhance trajectory prediction and generation.

\textbf{Analysis of Codebook Validity.} Furthermore, regarding codebook design, following similar previous work~\cite{chen2025enhancing,wang2025generative}, we analyze and verify that our codebook effectively captures geographical semantics from two critical perspectives: the spatial distribution of first-layer tokens across cities, and the semantic consistency of code combinations. 

\begin{figure}[h]
    \centering
    \includegraphics[width=0.95\linewidth]{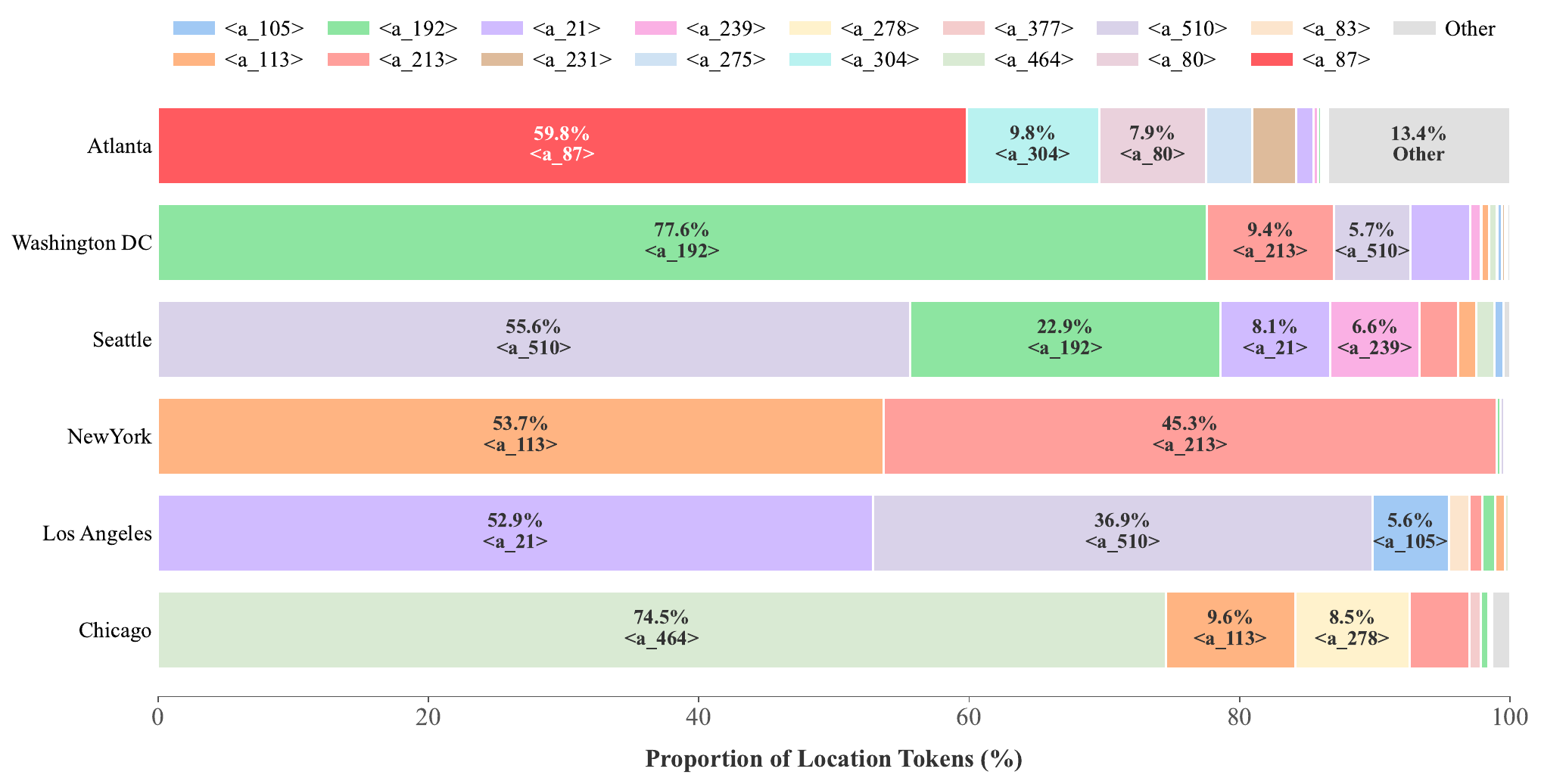} %
    \caption{\textbf{Distribution of First-Layer Tokens across Cities.} The visualizations show the proportion of location types (represented by first-layer tokens $<a_x>$) in each city. Different cities exhibit distinct dominant token distributions.}
    \label{fig:city_distribution}
\end{figure}

First, we visualize the distribution of first-layer tokens across different cities (Figure~\ref{fig:city_distribution}). We observe a significant variance in dominant tokens; for example, Chicago is primarily represented by token $<a_{464}><b_{x}><c_{x}><d_{x}>$ (74.5\%), whereas Los Angeles is dominated by $<a_{21}><b_{x}><c_{x}><d_{x}>$ (52.9\%). This demonstrates that the first layer successfully captures city-specific spatial characteristics and distinguishes between diverse urban environments.

\begin{figure}[htbp]
    \centering
    \subfloat[Semantic Profile Heatmap]{
        \includegraphics[width=0.90\linewidth]{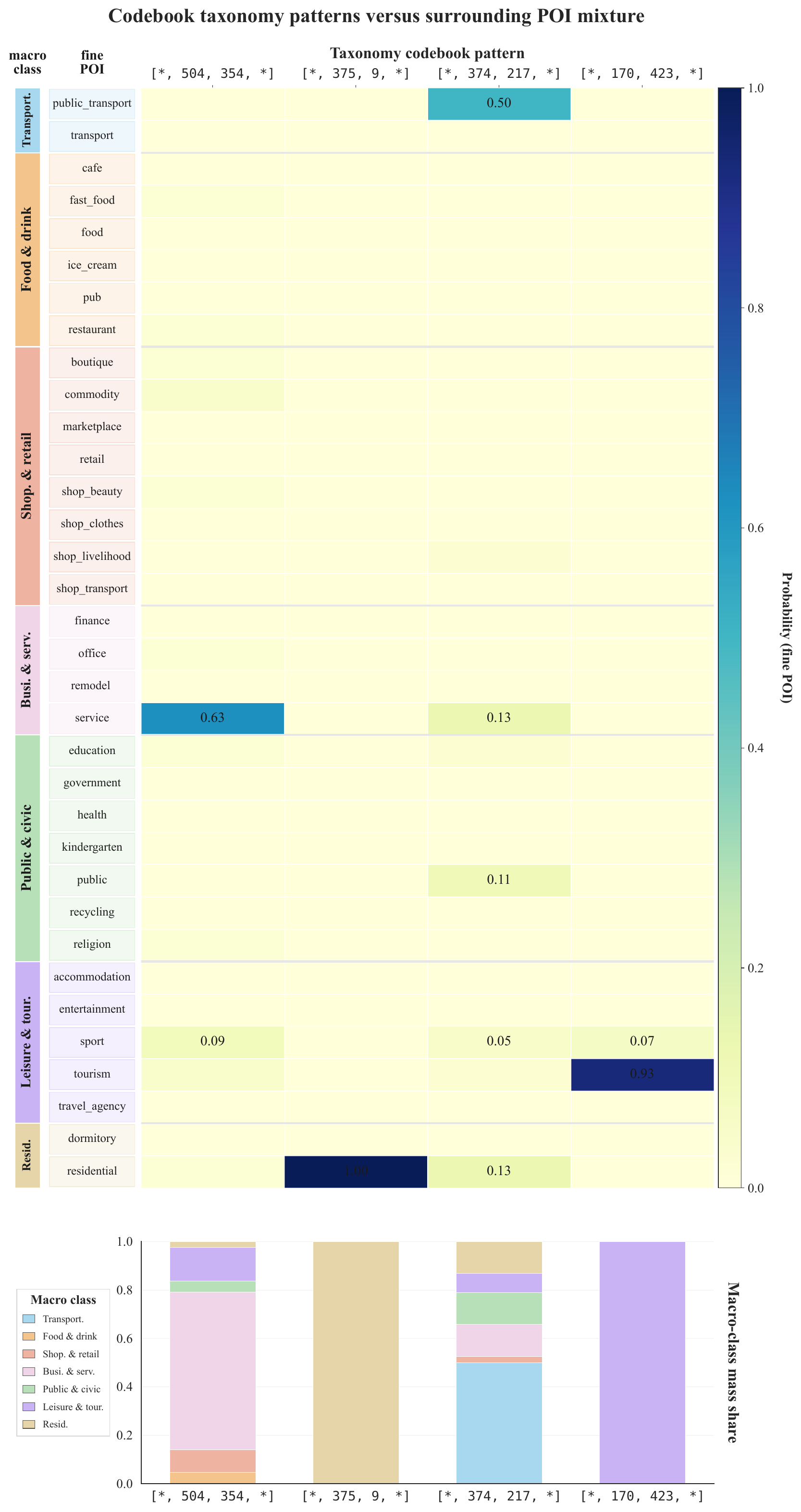}
        \label{fig:heatmap}
    }
    \\ 
    \subfloat[UMAP of POI Distributions]{
        \includegraphics[width=0.80\linewidth]{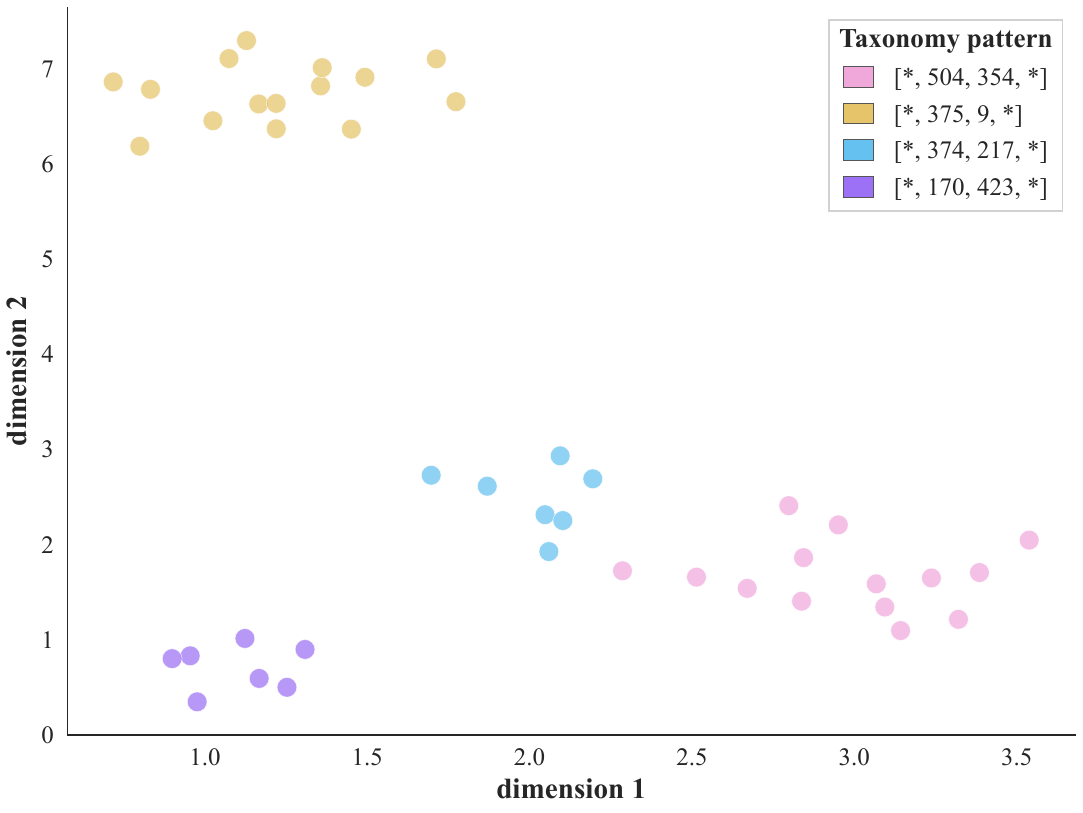}
        \label{fig:umap}
    }
    
    \caption{\textbf{Semantic Analysis of Code Combinations.} (a) The heatmap shows the average weight of POI categories for selected taxonomy codebook patterns. (b) UMAP visualization of the specific POI distribution vectors corresponding to these specific patterns. Distinct code combinations map to distinct semantic functions.}
    \label{fig:semantic_analysis}
\end{figure}

Second, we analyze the semantic consistency of specific code combinations using Points of Interest (POI) distributions. As shown in the UMAP visualization (Figure~\ref{fig:umap}), grids sharing the same taxonomy pattern (e.g., \texttt{[*, 504, 354, *]}, \texttt{[*, 375, 9, *]}, \texttt{[*, 374, 217, *]}, and \texttt{[*, 170, 423, *]}) form compact, well-separated clusters, indicating effective grouping of locations with similar functional attributes. Furthermore, the semantic profile heatmap (Figure~\ref{fig:heatmap}) explicitly reveals that distinct codes correspond to specific functional categories. For instance, the pattern \texttt{[*, 374, 217, *]} exhibits a high probability for ``Public Transport'' (0.50), while \texttt{[*, 375, 9, *]} is heavily dominated by ``Residential'' POIs (1.00). Similarly, \texttt{[*, 170, 423, *]} shows a strong mapping to ``Tourism'' (0.93), and \texttt{[*, 504, 354, *]} primarily captures ``Services'' functionality (0.63). These findings strictly confirm our codebook's ability to capture fine-grained semantic information, providing a robust foundation for conditional trajectory generation under novel constraints.

\begin{figure}[h]
    \centering
    \includegraphics[width=1.0\linewidth]{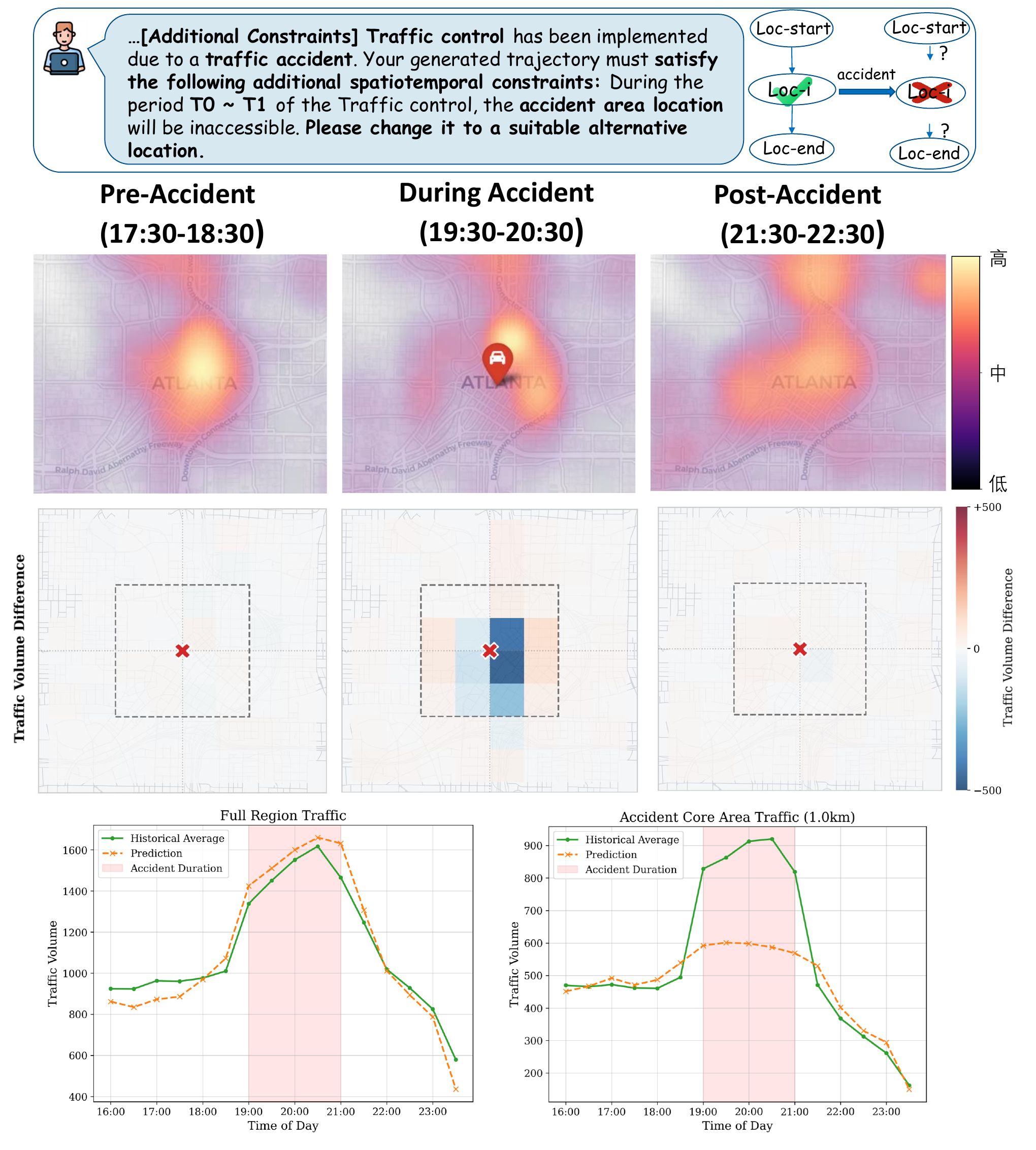}
    \vspace{-6mm}
    \caption{Spatiotemporal counterfactual simulation of a traffic accident in Atlanta. \textbf{Top:} The language prompt injecting spatiotemporal constraints and a schematic of the trajectory rerouting mechanism. \textbf{Row 1:} Traffic volume density heatmaps across three temporal phases: pre-accident (17:30--18:30), during the accident (19:30--20:30), and post-accident (21:30--22:30). \textbf{Row 2:} Spatial mapping of traffic volume differences between generated trajectories and historical averages, highlighting the volume drop (dark blue) at the accident center and the increase (red) in adjacent grids due to rerouting. \textbf{Row 3:} Temporal macroscopic and microscopic traffic volume comparisons over the full region (left) and the 1.0~km accident core area (right).}
    \vspace{-4mm}
    \label{fig:accident_simulation}
\end{figure}

\subsection{Case Study: Counterfactual Reasoning under Traffic Constraints}

To qualitatively validate MoveFM-R's capacity for conditional trajectory generation under novel constraints, we conducted a counterfactual simulation within a high-density urban area of Atlanta. Specifically, we designed an anomaly: a sudden traffic accident occurs at a central intersection at 19:00, resulting in localized traffic control measures that persist until 21:00. 
Under these spatiotemporal constraints, we synthesized the continuous travel trajectories of approximately 4,000 agents to evaluate the model's reasoning and dynamic rerouting capabilities. 

To establish a comparative baseline, we averaged the regional traffic flow from the preceding three days and contrasted it with our generated trajectories. Notably, this counterfactual scenario was introduced simply by appending the localized traffic constraints to the natural language prompt, successfully guiding the model to perform spatiotemporal avoidance reasoning without requiring any further fine-tuning or architectural modifications. Specifically, the following constraint is injected into the language input:

\begin{quote}
\itshape
[Additional Constraints] Traffic control has been implemented due to a traffic accident. Your generated trajectory must satisfy the following additional spatiotemporal constraints: During the period \texttt{\{CONTROL\_START\}}-\texttt{\{CONTROL\_END\}} of the traffic control, the accident area location \texttt{\{conflict\_str\}} will be inaccessible. Please change it to a suitable alternative location.
\end{quote}

This explicit instruction forces the Self-Reflective Mechanism to evaluate the compliance of the baseline trajectory against the new ``what-if'' condition. If a spatial violation is detected at the specified location token within the restricted timeframe, the model performs iterative \textit{Refinement} or \textit{Pruning} operations until the trajectory satisfies the constraint with minimal edits. Figure~\ref{fig:accident_simulation} visualizes the multi-scale spatiotemporal impact of this anomaly alongside the responsive trajectory generation of our model.

As illustrated in the first row of Figure~\ref{fig:accident_simulation}, the overall traffic density exhibits dynamic spatiotemporal shifts. Notably, during the active accident phase (middle panel, 19:30--20:30), the traffic density near the epicenter is substantially reduced compared to the pre-accident state. The second row precisely quantifies this visual reduction by mapping the spatial difference in traffic volume between the generated trajectories and historical averages. During the anomaly, a pronounced negative volume difference (dark blue) emerges at the accident center, directly corroborating the density drop observed in the first row. Concurrently, a corresponding increase in traffic volume (light red/orange) appears in the adjacent outer grids. This visually demonstrates the model's capacity to capture fine-grained spatial traffic redistribution, effectively simulating intelligent vehicular rerouting around restricted zones. In the pre- and post-accident phases, the difference map returns to near-zero values (white), confirming that in the absence of anomalies, the generated traffic seamlessly aligns with underlying normal patterns.

The line graphs (third row) further validate these temporal dynamics. The right panel focuses on the 1.0 km core area, where the generated traffic flow (orange dashed line) diverges with a significant decrease from the historical average (green solid line) precisely at 19:00, accurately mirroring the onset of traffic control measures. Following the event's conclusion at 21:00, the generated flow rapidly recovers to the historical baseline. Conversely, the left panel illustrates the macroscopic stability of the generated traffic across the broader region. Throughout the observation window, the overall generated flow remains remarkably consistent with the historical trend. This confirms that MoveFM-R successfully executes strict, localized risk aversion at the micro-level while preserving global macroscopic traffic demand, thereby avoiding unrealistic systemic collapse or distribution drift.

\section{Discussion and Conclusion}

In this work, we introduce MoveFM-R to bridge the fundamental gap between the statistical fidelity of Mobility Foundation Models (MFMs) and the semantic reasoning of Large Language Models (LLMs). Transcending simple concatenation, our approach achieves deep synergy through a structured alignment process that grounds continuous spatiotemporal coordinates in a shared semantic vocabulary. Furthermore, our novel self-reflective mechanism enables robust counterfactual generation via natural language, paving the way for more transparent, controllable, and interpretable urban mobility analysis.

\bibliographystyle{IEEEtran}
\bibliography{IEEEabrv,ref}

\begin{IEEEbiography}[{\includegraphics[width=1in,height=1.25in,clip,keepaspectratio]{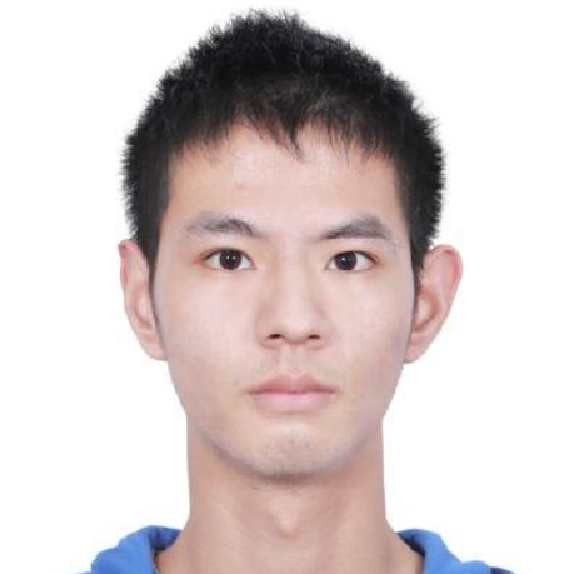}}]{Fanjin Meng} received the BS degree in the Department of Electronic Engineering from Tsinghua University, Beijing, China, in 2023. He is currently working toward the Master's degree in the Department of Electronic Engineering of Tsinghua University. His research interests include user behavior modeling and spatiotemporal data mining.
\end{IEEEbiography}

\vspace{-10mm}

\begin{IEEEbiography}[{\includegraphics[width=1in,height=1.25in,clip,keepaspectratio]{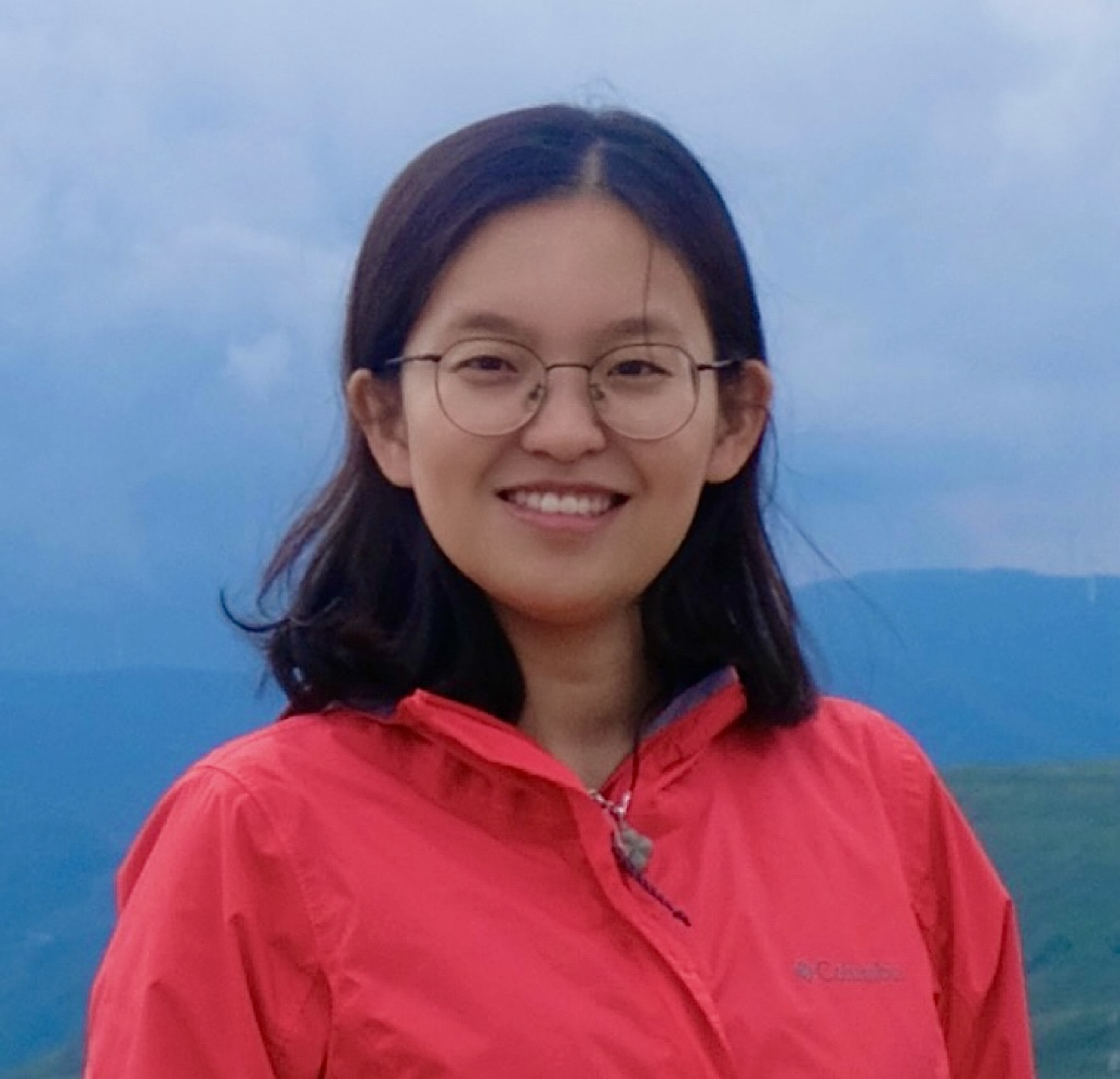}}]{Yuan Yuan} received the B.S. degree in electronic engineering from Tsinghua University, Beijing, China, in 2020, and the Ph.D. degree from the Department of Electronic Engineering, Tsinghua University, in 2025. She is currently a Postdoctoral Researcher with New York University (NYU). Her current research interests mainly focus on urban foundation models, urban simulations, and generative modeling of spatiotemporal systems. She has publications in conferences and journals such as ICLR, KDD, WWW, and IEEE Transactions on Knowledge and Data Engineering.
\end{IEEEbiography}
\vspace{-10mm}

\begin{IEEEbiography}[{\includegraphics[width=1in,height=1.25in,clip,keepaspectratio]{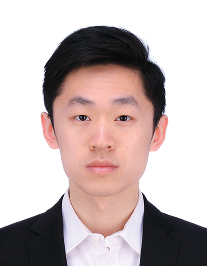}}]{Jingtao Ding} received the BS degrees in electronic engineering and the PhD degree in electronic engineering from Tsinghua University, Beijing, China, in 2015 and 2020, respectively. He is currently a post-doctoral research fellow with the Department of Electronic Engineering, Tsinghua University. His research interests include mobile computing, spatiotemporal data mining and user behavior modeling. He has more than 30 publications in journals and conferences such as IEEE Transactions on Knowledge and Data Engineering, ACM Transactions on Information Systems, KDD, NeurIPS, WWW, ICLR, SIGIR, IJCAI, etc.
\end{IEEEbiography}
\vspace{-10mm}

\begin{IEEEbiography}[{\includegraphics[width=1in,height=1.25in,clip,keepaspectratio]{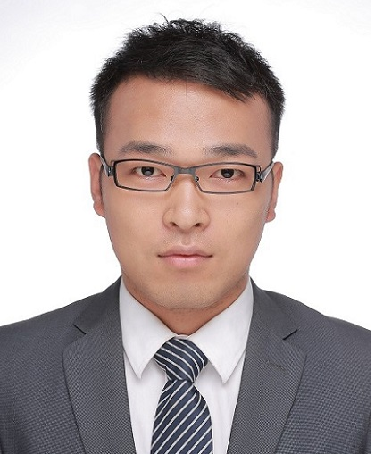}}]{Jie Feng} currently serves as a Researcher at Zhongguancun Academy. He received his B.E. and Ph.D. degrees from the Department of Electronic Engineering, Tsinghua University in 2016 and 2021. His research focuses on spatial intelligence, urban science, spatiotemporal data mining, large language models, and agents. He has published over 40 high-quality papers in prestigious conferences and journals such as KDD, NeurIPS, ICCV, and ACL. He was selected for the 2024 Stanford University list of the World's Top 2\% Scientists. Additionally, he serves as a Member of the Executive Committee of ACM SIGSPATIAL China and was honored as the 2025 ACM SIGSPATIAL China Rising Star Award in Spatial Data Intelligence.
\end{IEEEbiography}
\vspace{-10mm}

\begin{IEEEbiography}[{\includegraphics[width=1in,height=1.25in,clip,keepaspectratio]{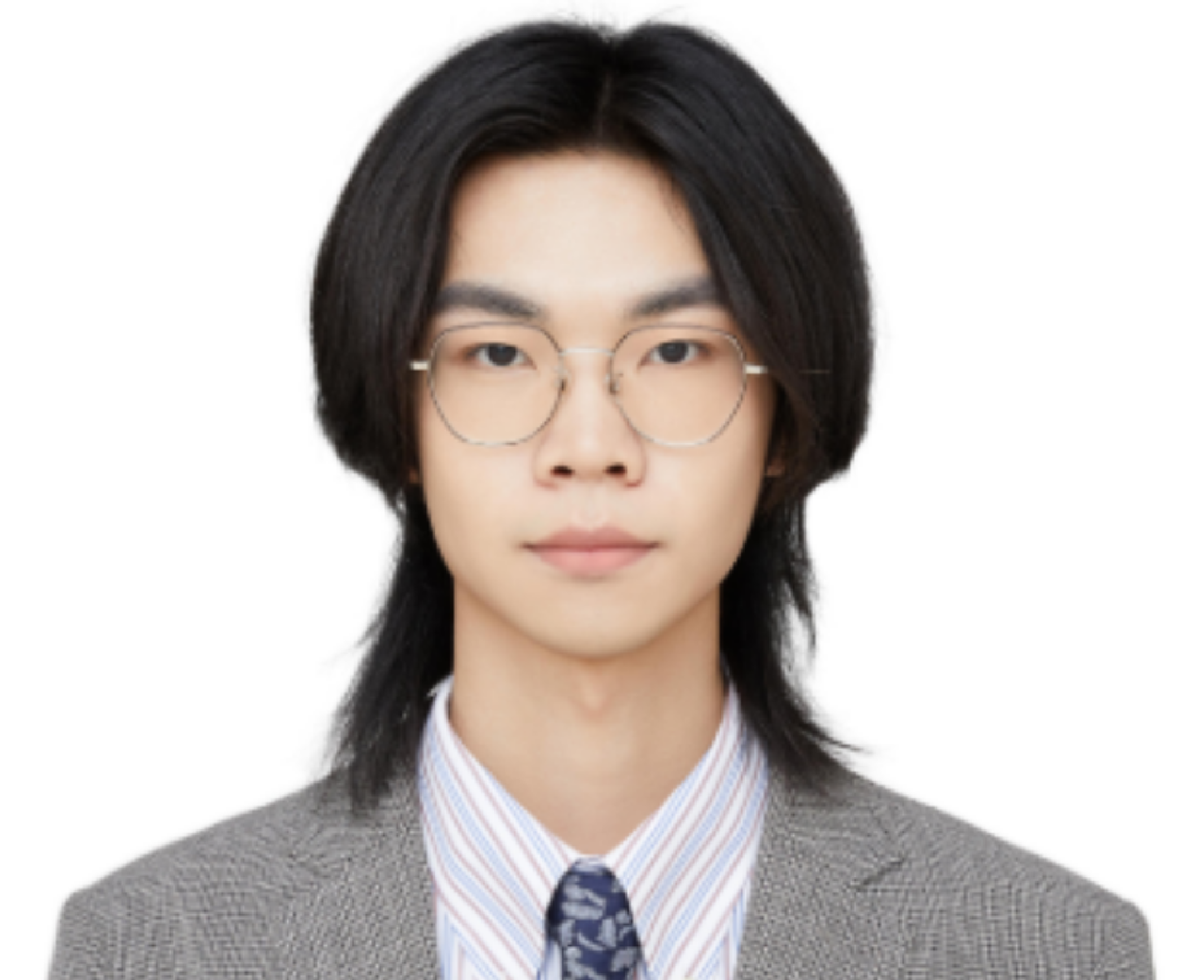}}]{Jie Feng} received the B.Eng. degree from the School of Mechatronical Engineering at Beijing Institute of Technology, Beijing, China. He is currently a Master's student in the Department of Electronic Engineering, Tsinghua University. His research interests include spatio‑temporal data mining and foundation models.
\end{IEEEbiography}
\vspace{-10mm}

\begin{IEEEbiography}[{\includegraphics[width=1in,height=1.25in,clip,keepaspectratio]{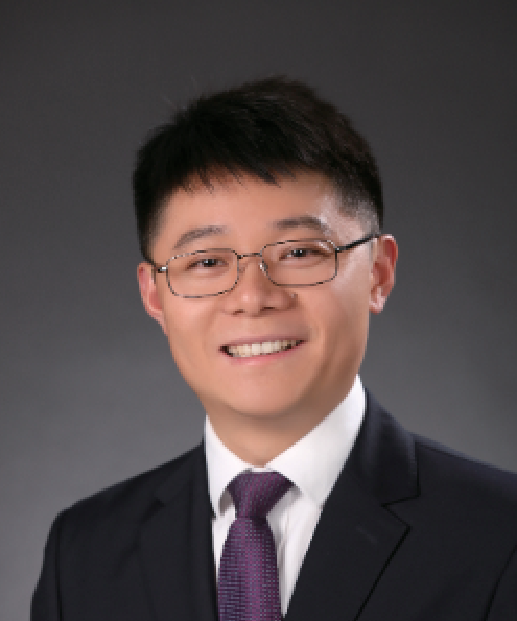}}]{Yong Li} (Member, IEEE) received the Ph.D. degree in signal and information processing from Beijing University of Posts and Telecommunications (BUPT), Beijing, China, in 2009. He is currently a Full Professor with the School of Information and Communication Engineering, BUPT. He has published more than 100 papers in journals, conference proceedings, and workshops, and filed over 60 patents. His current research interests include over-the-air testing, space-air–ground integrated networks, and beyond-5G systems. Dr. Li was a recipient of the First Grade Award of the Technological Invention from the China Institute of Communications.
\end{IEEEbiography}

\end{document}